\documentclass[10pt,twocolumn,letterpaper]{article}

\usepackage{cvpr}              

\usepackage[dvipsnames]{xcolor}
\usepackage{times}
\usepackage{epsfig}
\usepackage{graphicx}
\usepackage{amsmath}
\usepackage{amssymb}
\usepackage[accsupp]{axessibility}  

\usepackage{mathrsfs}
\usepackage{bm}
\usepackage{color}
\usepackage{threeparttable}
\usepackage{multirow}
\usepackage{algorithmic}
\usepackage{algorithm}

\usepackage[pagebackref,breaklinks,colorlinks]{hyperref}

\usepackage[capitalize]{cleveref}
\crefname{section}{Sec.}{Secs.}
\Crefname{section}{Section}{Sections}
\Crefname{table}{Table}{Tables}
\crefname{table}{Tab.}{Tabs.}

\def\cvprPaperID{6155} 
\def\confName{CVPR}
\def\confYear{2023}

\begin{document}

\title{Conditional Text Image Generation with Diffusion Models}

\author{
Yuanzhi Zhu,
Zhaohai Li, 
Tianwei Wang,
Mengchao He,
Cong Yao\footnotemark[2] \\
Alibaba DAMO Academy, Hangzhou, China. \\
\{z.yuanzhi, zhaohai.li, wangtw\}@foxmail.com, mengchao.hmc@alibaba-inc.com,\\
yaocong2010@gmail.com.
}
\maketitle

\renewcommand{\thefootnote}{\fnsymbol{footnote}} 
\footnotetext[2]{Corresponding author.} 

\begin{abstract}
Current text recognition systems, including those for handwritten scripts and scene text, have relied heavily on image synthesis and augmentation, since it is difficult to realize real-world complexity and diversity through collecting and annotating enough real text images. In this paper, we explore the problem of text image generation, by taking advantage of the powerful abilities of Diffusion Models in generating photo-realistic and diverse image samples with given conditions, and propose a method called \textbf{C}onditional \textbf{T}ext \textbf{I}mage \textbf{G}eneration with \textbf{D}iffusion \textbf{M}odels (CTIG-DM for short). To conform to the characteristics of text images, we devise three conditions: image condition, text condition, and style condition, which can be used to control the attributes, contents, and styles of the samples in the image generation process. Specifically, four text image generation modes, namely: (1) synthesis mode, (2) augmentation mode, (3) recovery mode, and (4) imitation mode, can be derived by combining and configuring these three conditions. Extensive experiments on both handwritten and scene text demonstrate that the proposed CTIG-DM is able to produce image samples that simulate real-world complexity and diversity, and thus can boost the performance of existing text recognizers. Besides, CTIG-DM shows its appealing potential in domain adaptation and generating images containing Out-Of-Vocabulary (OOV) words.
\end{abstract}

\section{Introduction}
\label{sec:intro}

Text recognition has been an important research topic in the computer vision community for a long time, due to its wide range of applications. In the past few years, numerous recognition methods for scene and handwritten text~\cite{shi2016crnn, shi2018aster, cluo2019moran, Deli2020srn, fang2021abinet, zhang2018multi, DAN_aaai20, Bhunia2021MetaHTR} have been proposed, which have substantially improved the recognition accuracy on various benchmarks. The volume and diversity of data are crucial for high recognition performance, but it is extremely hard, if not impossible, to collect and label sufficient real text images, so majority of the existing recognition methods rely heavily on data synthesis and augmentation.

\begin{figure}[t]
\centering
\vspace{-1mm}
\includegraphics[width=0.48\textwidth]{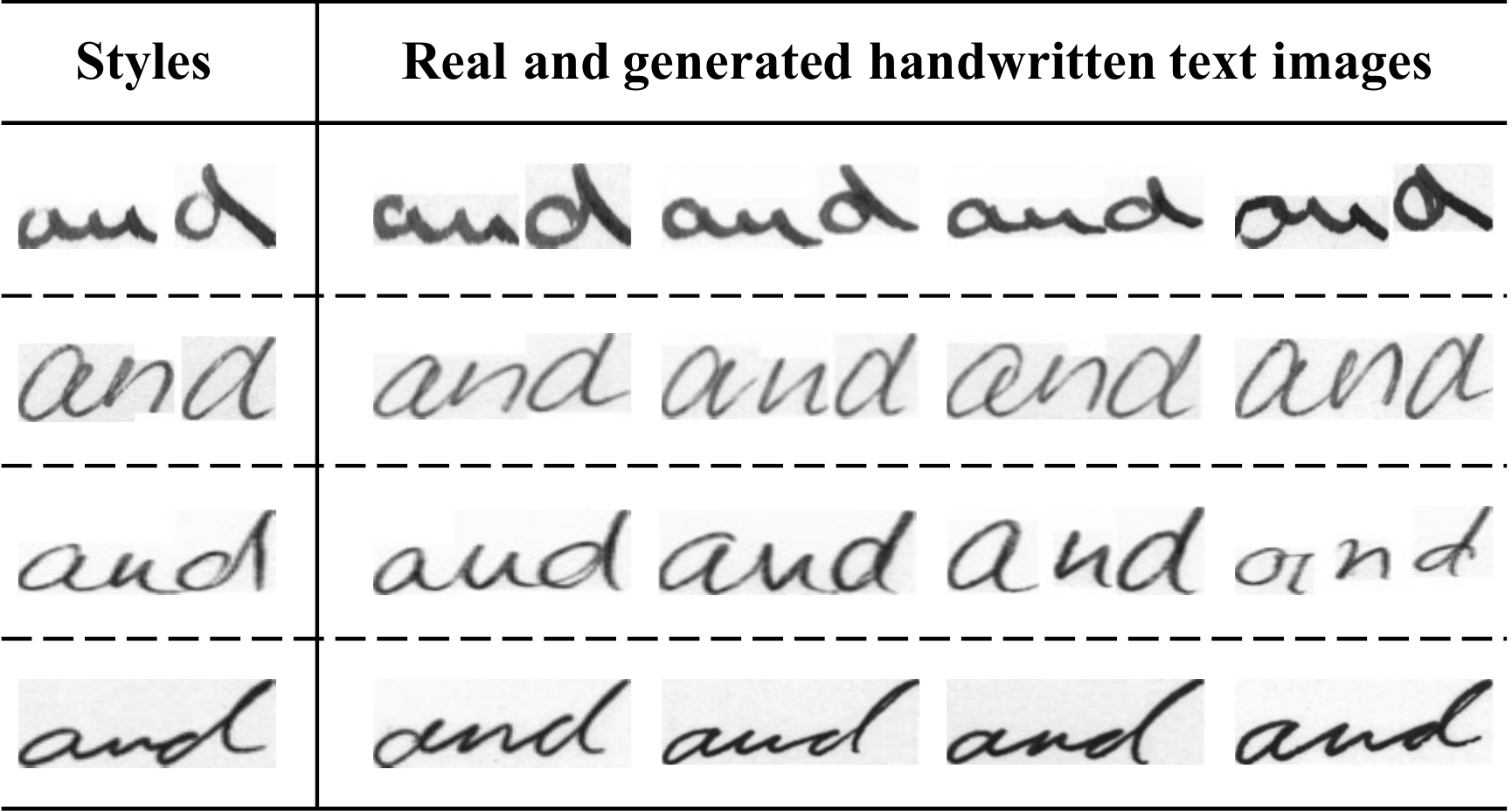}
\vspace{-4mm}
\caption{Handwritten text image samples from IAM~\cite{marti2002iam} or generated by our proposed CTIG-DM. On the left, the handwriting styles of the same word ``\textit{and}'' written by different writers are considerably different, indicating the diversity of handwritten text and the challenge of handwritten text recognition. On the right, two images out of four in each row are written by the corresponding writer on the left. Can you distinguish them from the generated samples? (The answer will be revealed in the next page.)}
\label{figure_introduction}
\vspace{-2mm}
\end{figure}

\begin{figure}[t]
\centering
\includegraphics[width=0.48\textwidth]{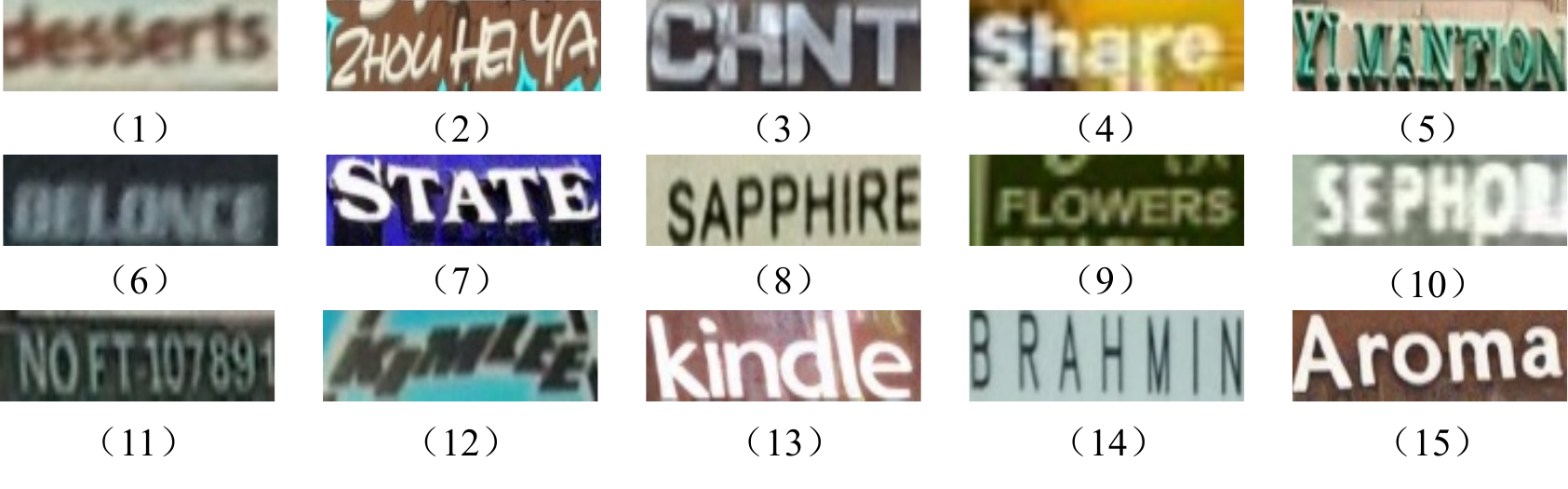}
\vspace{-6mm}
\caption{Scene text image samples from Real-L~\cite{Baek2021what} or produced by CTIG-DM. Only seven images herein are real. Can you identify them? (The answer will be revealed in the next page.)}
\label{figure_introduction_str}
\vspace{-4mm}
\end{figure}

Previously, a variety of data synthesis and augmentation methods~\cite{Jaderberg16mjsynth, gupta2016synthetic, long2020unrealtext, fogel2020scrabblegan, gan2021higan, Bhunia2021handt, luo2022slogan, kong2022look, Curtis2017data, luo2020learn} have been proposed to enrich data for training stronger text recognition models. In this paper, we investigate a technique that is highly related and complementary to such works. Drawing inspiration from the recent progress of Diffusion Models~\cite{alex2021improvedddpm, prafulla2021beats}, we propose a text image generation model, which is able to conduct data synthesis, and thus can boost the performance of existing text recognizers. 

A recent study~\cite{prafulla2021beats} has shown that State-Of-The-Art (SOTA) likelihood-based models\cite{alex2021improvedddpm} can outperform GAN-based methods~\cite{brock2018large, wu2019logan, karras2020analyzing} in generating images. 
Diffusion models~\cite{ho2020ddpm,alex2021improvedddpm,fran2015dm} have been becoming increasingly popular, due to their powerful generative ability in various vision tasks~\cite{Robin2022latentdiffusion,Chung2022come,Andreas2022repaint,omri2022blended,Jooyoung2021ilvr}. A typical representative of diffusion models is Denoising Diffusion Probabilistic Models (DDPM)~\cite{ho2020ddpm}. 
It generates diverse samples through different initial states of simple distribution and each transition. This means that it is challenging for DDPM to control the content of the output image due to the randomness of the initial states and transitions. Guided-Diffusion~\cite{prafulla2021beats} provides conditions to diffusion models by adding classifier guidance. UnCLIP~\cite{Aditya2022unclip} further pre-trains a CLIP model~\cite{radford2021clip} to match the image and whole text, which are used as the conditions for the diffusion models in image generation. 
While these approaches have focused on natural images, images with handwritten or scene text have their unique characteristics (as shown in Fig.~\ref{figure_introduction} and Fig.~\ref{figure_introduction_str}), which require not only \textit{image fidelity and diversity}, but also \textit{content validity} of the generated samples, i.e., the text contained in the images should be the \textit{same} as specified in the given conditions.

In this paper, we present a diffusion model based conditional text image generator, termed \textbf{C}onditional \textbf{T}ext \textbf{I}mage \textbf{G}eneration with \textbf{D}iffusion \textbf{M}odels (CTIG-DM for short). To the best of our knowledge, this is one of the first works to introduce diffusion models into the area of text image generation. The proposed CTIG-DM consists of a conditional encoder and a conditional diffusion model. Specifically, the conditional encoder generates three conditions, i.e., image condition, text condition, and style condition (the writing style of a specific writer). These conditions are proved to be critical for the fidelity and diversity of the generated text images. The conditional diffusion part uses these conditions to generate images from random Gaussian noise. 
As can be seen in Fig.~\ref{figure_introduction} and Fig.~\ref{figure_introduction_str}, the quality of the images generated by CTIG-DM is quite high that one can hardly tell them from real images\footnote{In Fig.~\ref{figure_introduction}, the real images are the first and last of each row. In Fig.~\ref{figure_introduction_str}, the real images are even numbered.}. By combining the given conditions, four image generation modes can be derived, i.e., synthesis mode, augmentation mode, recovery mode, and imitation mode. With these modes, various text images that can be used to effectively boost the accuracy of existing text recognizers (see Sec.~\ref{sec:experiments} for more details) could be produced. Moreover, CTIG-DM shows its potential in handling OOV image generation and domain adaptation.

The contributions can be summarized as follows: 
\begin{itemize}
\setlength{\itemsep}{0pt}
\item  We propose a text image generation method based on diffusion models, which is one of the first attempts to use diffusion models to generate text images.
\item We devise three conditions and four image generation modes, which can facilitate the generation of text images with high validity, fidelity, and diversity.
\item Experiments on both scene text and handwritten text demonstrate that CTIG-DM can significantly improve both the image quality and the performance of previous text recognizers. Besides, CTIG-DM is effective in OOV image generation and domain adaptation.
\end{itemize}

\section{Related Work}
\subsection{Text Recognition}
As an important task in computer vision, text recognition has attracted extensive attention in the community. 
Specifically, Scene Text Recognition (STR) and Handwritten Text Recognition (HTR) are the most popular research directions~\cite{Zhu2016SceneTD, Long2018SceneTD,chen2022textsurvey}.

Scene text images generally contain complex backgrounds and irregular text arrangements. 
Early, He et al.~\cite{he2016reading} and Shi et al.~\cite{shi2016crnn} proposed to model STR as a sequence-to-sequence mapping issue by combining CNN, RNN, and CTC~\cite{Alex2006ctc}. 
Then, attention-based methods~\cite{shi2018aster,cluo2019moran,DAN_aaai20} gradually emerged and achieved a breakthrough in irregular text recognition. 
In recent years, benefiting from the success of Transformer~\cite{Ashish2017transformer}, many methods~\cite{Deli2020srn,fang2021abinet,da2022levocr,wang2022multi} improved the recognizer from the perspective of the language model.

Handwritten text images have diverse writing styles and difficult-to-segment cursive joins. 
Zhang et al.~\cite{zhang2019sequence} addressed the handwriting style diversity problem by domain adaption. 
Bhunia1 et al.~\cite{Bhunia2021MetaHTR} employ Model Agnostic meta-learning algorithm to train writer adaptive HTR network. 
Recently, due to the lack of real data, more researchers paid attention to the fields of text data augmentation and synthesis, thereby improving the performance of handwritten text recognizers~\cite{alonso2019adver,Kang2019unsupervised,bhunia2019low,fogel2020scrabblegan,luo2020learn,luo2022slogan}.

\begin{figure*}[t]
\centering
\vspace{-3mm}
\includegraphics[width=0.9\textwidth]{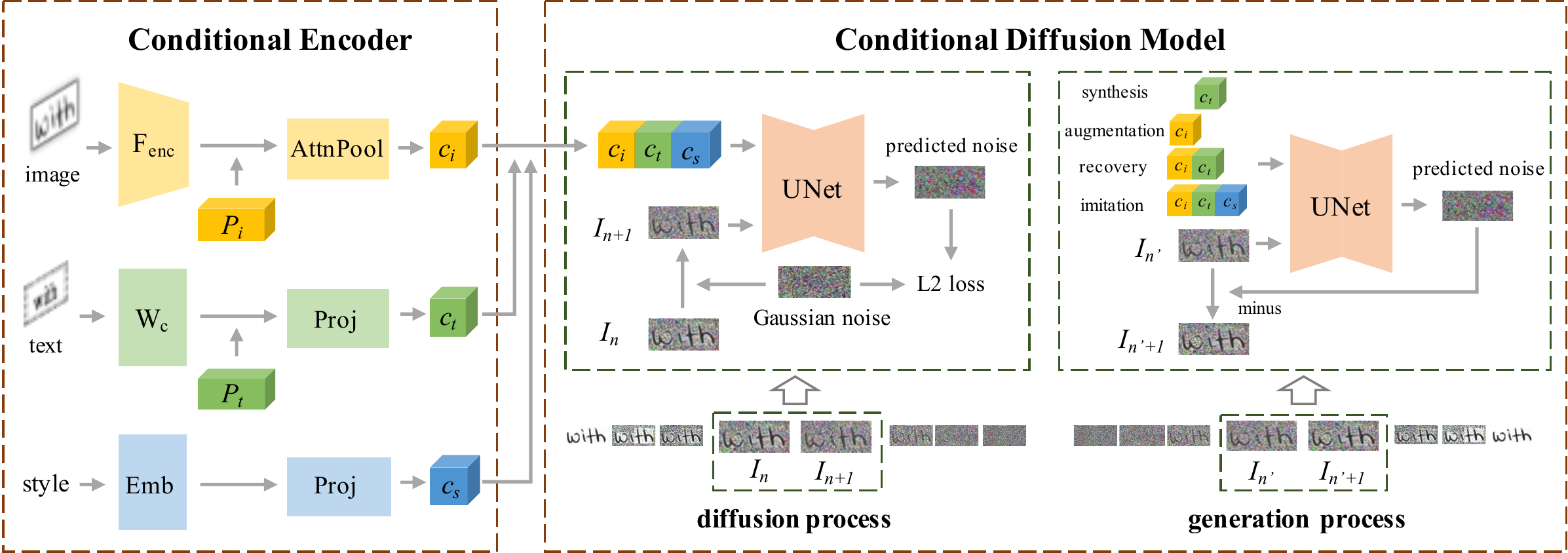}
\vspace{-2mm}
\caption{The overall architecture of the proposed CTIG-DM.}
\vspace{-4mm}
\label{figure_pipline}
\end{figure*}

\subsection{Text Image Augmentation and Synthesis}
Wigington et al.~\cite{Curtis2017data} and Bhunia et al.~\cite{bhunia2019low} built grids on the original images and hidden features, respectively, and then augmented them by adding random perturbations. 
Luo et al.~\cite{luo2020learn} proposed a learnable augmentation method to obtain more controllable samples. 
These methods have made significant progress in improving the performance of text recognizers. 
However, they fail to create OOV samples and thus the diversity is limited by the training set. 

There are many GAN-based handwritten text image synthesis methods. 
Fogel et al.~\cite{fogel2020scrabblegan} presented a semi-supervised approach that can generate images of words with a variable length. 
Kang et al.~\cite{Kang2019unsupervised} generated credible handwritten word images by adjusting calligraphic style features and textual contents. 
Luo et al.~\cite{luo2022slogan} proposed a style bank and dual discriminators to solve the issues of style representation and content embedding. 
However, GAN-based methods are generally difficult to train and require careful design of hyper-parameters, otherwise, it is easy to fall into mode collapse~\cite{prafulla2021beats}. 
Besides, most of the above data synthesis methods are only experimented on handwritten Latin text without exploring generalization to other types of texts, such as scene text or handwritten Chinese text. 

\subsection{Diffusion Models}
Recently, a category of deep generative models, named diffusion models~\cite{fran2015dm}, has achieved impressive results in computer vision tasks, outperforming GAN-based methods in the diversity of generated images~\cite{prafulla2021beats}. 
Inspired by the non-equilibrium thermodynamics theory~\cite{fran2015dm}, Ho et al.~\cite{ho2020ddpm} learned to model the Markov transition from noise to data distribution, enabling unconditional image generation. 
Luhman et al.~\cite{Luhman2020dmhand} focused on online handwriting and generated point sequences based on diffusion models. 
Prafulla et al.~\cite{prafulla2021beats} introduced additional classifiers to provide conditions for the diffusion models. 
Nichol et al.~\cite{Alexander2022glide} explored diffusion models for text-conditional image synthesis with classifier-free guidance.
Ramesh et al.~\cite{Aditya2022unclip} proposed a two-stage model, i.e., a prior that generates a CLIP image embedding given a text caption, and a decoder that generates an image conditioned on the image embedding.

Unlike general image generation, text image generation requires more unique contexts and textural features at the character-level.  
This inspires us to use a text recognizer to obtain conditions related to text images. 

\section{Methodology}
\subsection{Conditional Encoder}
As illustrated in Fig.~\ref{figure_pipline}, the outputs of the conditional encoder consist of an image condition $c_i$, a text condition $c_t$, and a style condition $c_s$. 
Specifically, $c_i$ represents the unique visual characteristics of input images. 
$c_t$ indicates the semantic characteristics of input contexts. 
$c_s$ describes the information about the writer styles, which are only used in the handwritten text generation. 

\textbf{Image condition.} 
Different from the patterns of natural images, the rich visual information of text images concentrates on the characters. 
Therefore, we propose to use the encoder of a pre-trained text recognizer to obtain image condition that can better express \textit{general features}~(e.g., textural and colors) rather than the noisy features~(e.g. the backgrounds). 
The background information comes from the training data and is encoded in the trained diffusion models.
Given the input image $I$, the generation process of the image condition can be formulated as
\begin{align}
c_i=AttnPool(F_{enc}(I) + Emb(P_i)),
\end{align}
where $F_{enc}$ is the feature extractor of the pre-trained text recognizer.
$P_i$ denotes the index of encoded image patches, and $Emb$ is the embedding function which encodes $P_i$ to obtain position embedding. 
$AttnPool$ represents the attention pooling~\cite{Lee2019attnpool} to customize visual representations. 
The image condition is an image-level representation, which is an aggregation of patch-level features.

\textbf{Text condition.}
Text condition specifies the \textit{contents} of the generated text images and represents the unique contexts among chars, which is critical for the proposed CTIG-DM. 
Although the pre-trained word embedding~\cite{Devlin2019bert} is widely used in the natural language processing field, it cannot handle the text image generation of OOV words. 
Therefore, we adapt the classifier weights $W_c$ of the pre-trained text recognizer described above to encode text condition. 
The generation process of text condition can be described as
\begin{align}
c_t=Proj(W_c T + Emb(P_t)), 
\end{align}
where $T$ and $P_t$ represent the one-hot encoding and the index of characters in the text string label. 
A linear projection layer $Proj$ is applied to unify the dimensionality with the text condition. 

\textbf{Style condition.}
The style condition is particularly designed for text generation of HTR, which contains the writing characteristics (e.g., character slants, cursive joins, and stroke widths) of a specific writer.  
In other words, the style condition represents \textit{personal style} rather than image style.
The generation of style condition can be represented as
\begin{align}
c_s=Proj(Emb(S)), 
\end{align}
where $S$ is the writer ID. Overall, the combination of $c_i$, $c_t$, and $c_s$ is fed into the conditional diffusion model to generate text images. 

\subsection{Conditional Diffusion Model}
Different from natural image generation in vanilla diffusion models~\cite{ho2020ddpm,alex2021improvedddpm,prafulla2021beats}, the proposed CTIG-DM introduces expert knowledge related to text image, i.e., image condition, text condition, and style condition. 
As shown in Fig.~\ref{figure_pipline}, following~\cite{ho2020ddpm}, the conditional diffusion model is implemented by UNet~\cite{Ronneberger2015unet}, which contains two processes, i.e., the diffusion process and the generation process. 

\textbf{Diffusion process.} 
In the diffusion process, by continuously adding Gaussian noise to an initial image, the characteristics of the initial image will gradually disappear, and the image eventually becomes standard Gaussian noise. 
Specifically, as illustrated in Fig.~\ref{figure_pipline}, at the diffusion step $n$, the next noisy image $I_{n+1}$ can be obtained by
\begin{align}
I_{n+1} = I_n + \epsilon, 
\end{align}
where $\epsilon$ denotes the Gaussian noise.  
Then the next noisy image and the proposed conditions are fed into the UNet to predict the distribution of the noise, which is supervised by the squared error loss 
\begin{align}
L=||\epsilon - \epsilon_{\theta} ([c_i, c_t, c_s], I_{n+1})||^2. 
\end{align}
$\epsilon_{\theta}$ and $[\cdot]$ represent the parameters of UNet and the concatenation process respectively. 
Specifically, given the diffusion step, $L$ can be directly calculated by the original image rather than the intermediate status~\cite{ho2020ddpm}. 

\textbf{Generation process.}
\label{generation_process}
During the generation process, the image is finally generated by gradually denoising the random initial Gaussian noise.
Specifically, at the generation step $n'$, the current denoising image $I_{n'}$ and proposed conditions are fed into UNet to predict the noise.
Then the next denoising image $I_{n'+1}$ is the minus between predicted noise and the current denoising image, as follows,
\begin{align}
I_{n'+1} = I_{n'} - \epsilon_{\theta} (\textbf{c}, I_{n'}).
\end{align}
\textbf{c} stands for the different settings of proposed conditions.
Benefiting from the input of various conditions, we propose four image generation modes, i.e., synthesis mode, augmentation mode, recovery mode, and imitation mode. 
Tab.~\ref{synthesis_mode} shows the details for each generation mode. 
In empirical experiments, the text condition determines the diversity and is helpful for OOV image generation, while the image condition and style condition influence the fidelity. 
More details can be found in Sec.~\ref{section_ablation}.

\begin{table}[htbp]
\vspace{-1mm}
  \centering
  \caption{The combination of conditions for each generation mode.}
  \vspace{-1mm}
    \begin{tabular}{cccc}
    \hline
    \textbf{Generation Modes}      & $c_{i}$    & $c_{t}$     & $c_{s}$ \\
    \hline
    Synthesis &       & \checkmark      &  \\
    Augmentation & \checkmark      &       &  \\
    Recovery & \checkmark      & \checkmark      &  \\
    Imitation  & \checkmark   & \checkmark      & \checkmark \\
    \hline
    \end{tabular}%
  \label{synthesis_mode}%
\vspace{-3mm}  
\end{table}%

\section{Experiments} 
\label{sec:experiments}
\subsection{Datasets}
\textbf{Handwritten text datasets.}
IAM~\cite{marti2002iam} contains more than 115,000 words written in English by 657 different writers. 
RIMES~\cite{Grosicki2009rimes} contains more than 60,000 words written in French by over 1,000 authors. 
IAM and RIMES are widely used in previous methods~\cite{Jorge2018offline,fogel2020scrabblegan,bhunia2019low,luo2020learn,luo2022slogan} and can serve for a variety of handwritten recognition tasks. 
CVL~\cite{Kleber2013CVL} contains seven different handwritten texts (one in German and six in English) written by 311 different writers. 
We use the English part for the experiment of domain adaptation. 
CASIA-HWDB 1.0-1.1~\cite{liu2011casia} consists of 2,678,424 images of offline handwritten Chinese characters. 
We use it for OOV handwritten Chinese characters generation. 

\textbf{Scene text datasets.}
We use MJSynth~\cite{Jaderberg2014mj,Jaderberg16mjsynth}, SynthText~\cite{gupta2016synthetic}, and Real-L~\cite{Baek2021what}, as training data. 
The test datasets consist of regular datasets, i.e., IIIT 5K-Words (IIIT)~\cite{Mishra2012iiit5k}, Street View Text (SVT)~\cite{Wang2011svt}, ICDAR 2003 (IC03)~\cite{Simon2003ic03} and ICDAR 2013 (IC13)~\cite{Dimosthenis2013ic13}, and irregular datasets i.e., ICDAR 2015 (IC15)~\cite{Dimosthenis2015ic15}, Street View Text-Perspective (SVTP)~\cite{Trung2013svtp} and CUTE80 (CUTE)~\cite{Anhar2014cute}.
Details of these datasets can be found in previous works~\cite{Deli2020srn}. 

\begin{figure*}[t]
\centering
\vspace{-3mm}
\includegraphics[width=0.98\textwidth]{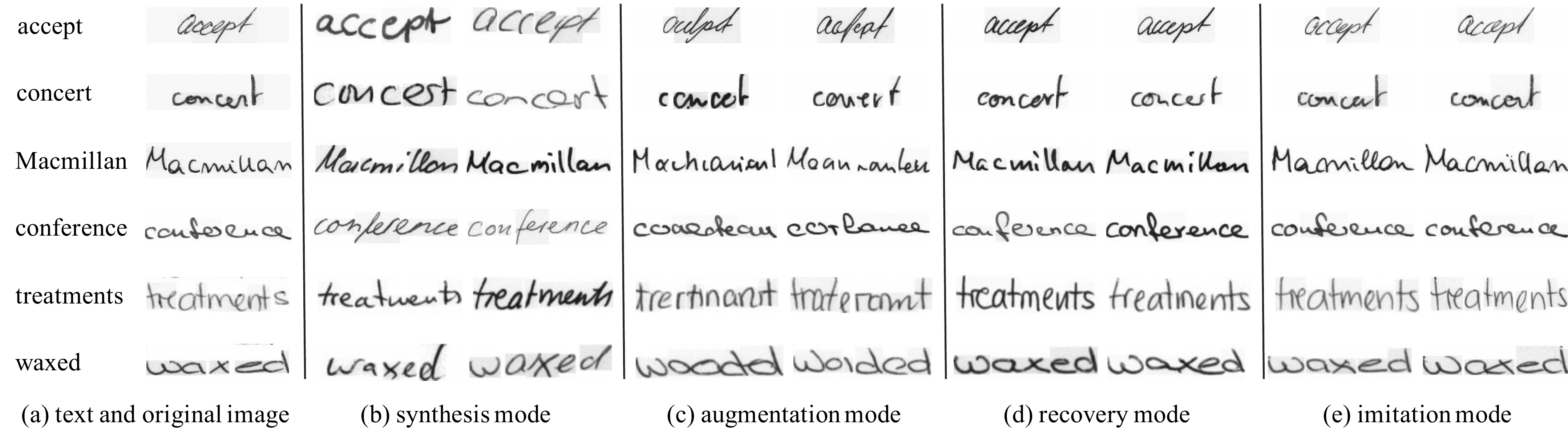}
\vspace{-3mm}
\caption{Visualization of handwritten text images and their text strings and different generation modes. }
\vspace{-5mm}
\label{figure_ablation}
\end{figure*}

\subsection{Implementation Details}
\textbf{Handwritten text. }
Similar to previous work~\cite{fogel2020scrabblegan,luo2022slogan} on the IAM, RIMES, and CVL datasets, the height of the training images is resized to 64 pix and the width is calculated with the original aspect ratio (up to 256 pix). 
The evaluation criteria of recognition performance are Word Error Rate (WER \%) and Character Error Rate (CER \%). 
The WER respects the ratio of the error at the word level, and the CER corresponds to the edit distance between the recognition result and ground-truth, normalized by the length of ground-truth. 
With respect to the generation quality, FID~\cite{Martin2017fid}, GS~\cite{Valentin2018gs}, SSIM, RMSE and LPIPS~\cite{Zhang2018lpips} are introduced to our experiments. 
Lower values of WER, CER, FID, GS, RMSE and LPIPS, and higher value of SSIM are preferable.

\textbf{Scene text.}
When combined with previous methods, all experimental settings are kept the same for the sake of fair comparison, except that generated data from the proposed CTIG-DM is used. 

\textbf{Network.}
The pre-train text recognizer uses CRNN architecture~\cite{shi2016crnn} and the diffusion models follow the DDPM architecture~\cite{ho2020ddpm,alex2021improvedddpm}. 
Specifically, the proposed conditions, whose dimensionality is set to 512, are concatenated to the time-step embedding described in GLIDE~\cite{Alexander2022glide}. 
To avoid the proposed CTIG-DM over-relying on any condition, we randomly set the $c_i$, $c_t$, and $c_s$ to a learnable embedding 20\%, 10\%, and 20\%, respectively, of the time. 

\textbf{Optimization.}
We use the AdamW~\cite{Ilya2019adam} as the optimizer with the settings of $\beta_1$ = 0.9, $\beta_2$ = 0.999, and $weight\_decay$ = 0.2. 
The learning rate is 0.0001 with cosine annealing. 
The batch size is set to 64 for training recognizer and 256 for training diffusion models. 
All experiments are conducted on NVIDIA Tesla V100 GPUs.

\begin{figure*}[h]
\centering
\vspace{-3mm}
\includegraphics[width=0.98\textwidth]{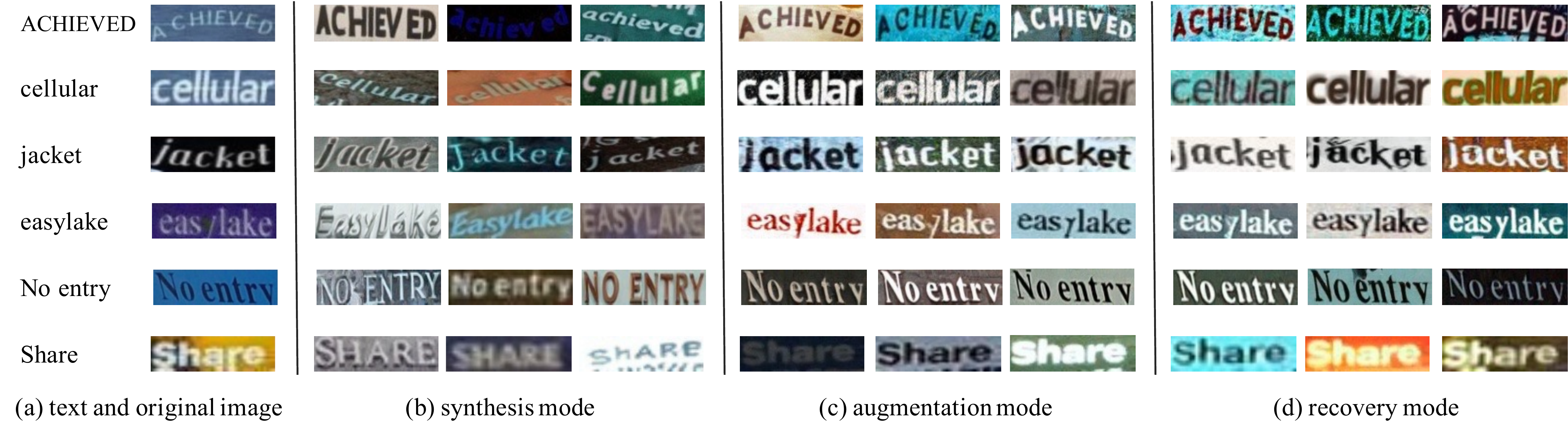}
\vspace{-3mm}
\caption{Visualization of scene text images and their text strings and different generation modes.}
\vspace{-4mm}
\label{figure_ablation_str}
\end{figure*}

\subsection{Ablation Study}
\label{section_ablation}
In this subsection, we conduct ablation study on the IAM dataset. Concretely, we explore the role of various conditions obtained by the conditional encoder in the diffusion process and show visualizations of each generation mode during the generation process. 

As presented in Tab.~\ref{ablation_condition}, the trends of different metrics are consistent.  
First, the baseline is built without any conditions, which performs a large FID score of 33.42. 
Then, we add $c_{i}$ and $c_{t}$ respectively, and find that both FID scores are significantly improved, which indicates that the image and text features have critical supports for text image generation. 
Besides, by comparing the above two experiments, we find that $c_{t}$ has a greater impact on the quality of image generation. 
When both $c_{i}$ and $c_{t}$ are added, the FID further decreases to 9.76. 
Moreover, we add $c_{s}$ to further guide the proposed CTIG-DM to present specific handwriting styles that can be easily identified. 
Finally, with all conditions, our method achieves the best FID of 9.34.

\begin{table}[htbp]
\small
\centering
\vspace{-1mm}
\caption{Effectiveness of different conditions in diffusion process. } 
\vspace{-2mm}
\begin{tabular}{cccccccc}
\hline
$c_{i}$ & $c_{t}$ & $c_{s}$ & \textbf{FID} $\downarrow$ & \textbf{SSIM} $\uparrow$ & \textbf{RMSE} $\downarrow$ & \textbf{LPIPS} $\downarrow$ \\
\hline
      &       &       & 33.42 & 0.5346 & 0.2195 & 0.4144 \\

\checkmark  &       &       & 13.36 & 0.6476  & 0.1994 & 0.2337 \\

      & \checkmark &       & 10.92 & 0.7077 & 0.1738  &  0.1349 \\

\checkmark  & \checkmark  &       & 9.76 & 0.7241  & 0.1697  & 0.1132 \\

\checkmark  & \checkmark  & \checkmark     & 9.34 & 0.7374 & 0.1501  & 0.0936 \\
\hline
\end{tabular}%
\label{ablation_condition}%
\vspace{-2mm}
\end{table}%

Benefiting from the input of diverse conditions, Fig.~\ref{figure_ablation} and Fig.~\ref{figure_ablation_str} illustrate the image visualizations of different generation modes for handwritten text and scene text, respectively. 
Specifically, since the synthesis mode only depends on the $c_t$, we observe that its generated images have rich diversity. 
As presented in Fig.~\ref{figure_ablation}~(b) and Fig.~\ref{figure_ablation_str}~(b), handwritten text images exist in a variety of character slants, ink blots, cursive joins, stroke widths, and paper backgrounds, while scene text images are diverse in text rotations, backgrounds, blur noise, and fonts. 
Correspondingly, the $c_i$ is more critical for the fidelity of the generated images. 
In augmentation mode, we find that the generated images are similar in overall appearance to the original images, but part details are lost in the specific characters. 
As shown in Fig.~\ref{figure_ablation}~(c) and Fig.~\ref{figure_ablation_str}~(c), images are more likely to generate wrong characters when the $c_i$ is used solely. 
This is because only the image attributes are included in $c_i$ and the unique contexts among characters included in $c_t$ are missing. 
After adding $c_t$, the generated images in Fig.~\ref{figure_ablation}~(d) and Fig.~\ref{figure_ablation_str}~(d) rarely have wrong characters and contain diversity at the same time. 
Finally, when $c_s$ designed for handwritten text is added, the fidelity of the generated images in Fig.~\ref{figure_ablation}~(e) improves further, indicating that $c_s$ plays an important role in image generation. 
In Sec.~\ref{section_oov}, we will demonstrate the effect of $c_s$ on the style control of generated text images. 
Since the synthesis mode and the imitation mode (recovery mode in scene text) have the best diversity and fidelity, respectively, the generated data (denoted as MIX) we use in Sec.~\ref{recognition_exp} are derived from an equal-proportion mixture of these two modes.

\begin{figure}[h]
\centering
\includegraphics[width=0.48\textwidth]{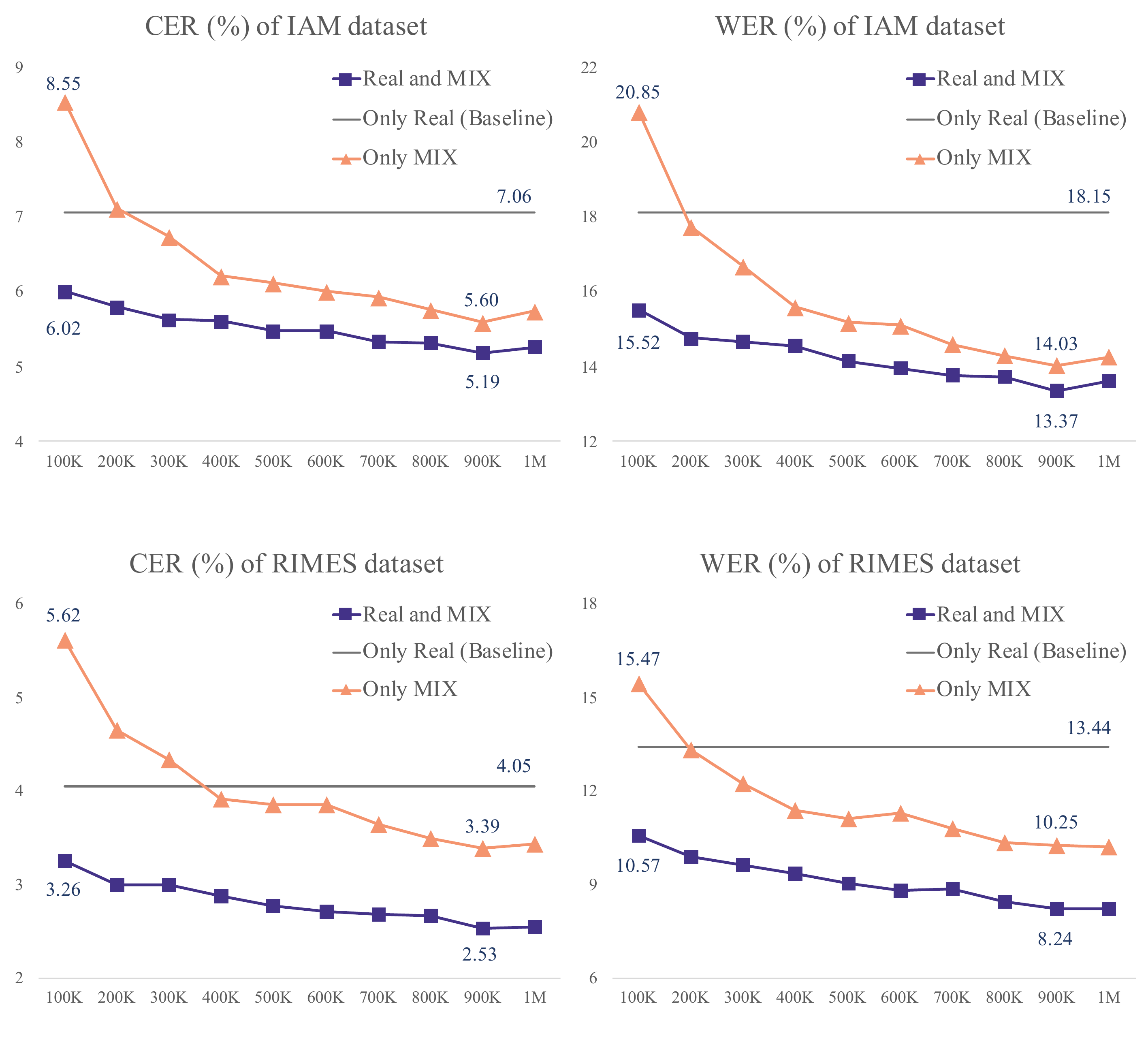}
\vspace{-4mm}
\caption{Effectiveness of our generated data (denoted as MIX) on IAM and RIMES (denoted as Real). The horizontal and vertical axes represent the amount of MIX used and the recognition performance (lower values are preferable), respectively.}
\vspace{-5mm}
\label{figure_amount}
\end{figure}

\subsection{Recognition Performance}
\label{recognition_exp}
For text recognition tasks, the ultimate purpose of generating images is to augment the training set and improve the performance of the recognizers. 
Therefore, we conduct experiments on multiple types of text recognition, including English handwritten text recognition, French handwritten text recognition, and scene text recognition, to demonstrate the validity of the generated data on the recognizers.

\subsubsection{Handwritten Text Recognition}
Following the settings of previous works~\cite{bhunia2019low,luo2020learn,luo2022slogan}, we use CRNN~\cite{shi2016crnn} as the recognizer. 
Fig.~\ref{figure_amount} illustrates the effectiveness of our generated data. 
The horizontal line on each subfigure indicates the baseline where the real training set is used solely. 
Specifically, for IAM dataset, about 200K generated data can achieve comparable performance as the baseline. 
Simultaneously, as the amount of MIX gradually increases, CER and WER are steadily decreasing, and realize the best results, i.e, CER of 5.60\% and WER of 14.03\%, at 900K. 
The experiment results on RIMES dataset show similar trends. 
Moreover, as described in the purple line in Fig.~\ref{figure_amount}, the recognizer performance can be further improved when using both real and generated data. 
Finally, compared with the baseline, on the IAM dataset, WER and CER decrease by 4.78\% and 1.87\%, respectively, and on the RIMES dataset, WER and CER decrease by 5.20\% and 1.52\%, respectively. 
This indicates that in HTR with a limited real training set, the generated data of our method can significantly improve the recognizer performance, which demonstrates the validity of the proposed CTIG-DM.

\begin{table}[h]
\small
  \centering
  \caption{Comparison of recognition performance with previous methods on the IAM and RIMES datasets. The numbers of ``$\Delta$'' denote the improvements from the baseline to our method. ``Aug*'' represents the random geometric augmentation of~\cite{luo2020learn}.}
    \begin{tabular}{cccccc}
    \hline
    \multirow{2}[4]{*}{} & \multicolumn{2}{c}{\textbf{IAM}} &       & \multicolumn{2}{c}{\textbf{RIMES}} \\
\cline{2-3}\cline{5-6}          & \textbf{WER}   & \textbf{CER}   &       & \textbf{WER}   & \textbf{CER} \\
    \hline
    Sueiras et al.~\cite{Jorge2018offline}    & 23.80  & 8.80   &       & 15.90  & 4.80 \\
    Alonso et al.~\cite{alonso2019adver}     & -     & -     &       & 11.90  & 4.03 \\
    Zhang et al.~\cite{zhang2019sequence}    & 22.20  & 8.50   &       & -     & - \\
    Bhunia et al.~\cite{bhunia2019low}    & 17.19 & 8.41  &       & 10.47 & 6.44 \\
    ScrabbleGAN~\cite{fogel2020scrabblegan} & 23.61 & 13.42 &       & 11.32 & 3.57 \\
    Kang et al.~\cite{Kang2019unsupervised}    & 17.26 & 6.75  &       & -     & - \\
    Luo et al.~\cite{luo2020learn} & 14.04 & 5.34 & & 9.23 & 2.57 \\
    SLOGAN~\cite{luo2022slogan} & 14.97 & 5.95  &       & 11.50  & 3.35 \\
    \hline
    Baseline  & 18.15 & 7.06 &  & 13.44 & 4.05 \\    
    Ours & \textbf{13.37} & \textbf{5.19}  &       & \textbf{8.24}  & \textbf{2.53} \\
    $\Delta$ & \textcolor{Green}{\textbf{+4.78}} & \textcolor{Green}{\textbf{+1.87}} & & \textcolor{Green}{\textbf{+5.20}} &  \textcolor{Green}{\textbf{+1.52}} \\ 
    Ours + Aug*  & \underline{12.01} & \underline{4.67}  &       & \underline{6.89}  & \underline{1.98} \\
    \hline
    \end{tabular}%
  \label{recognition_preformance}
  \vspace{-4mm}
\end{table}%

\begin{table*}[t]
  \centering
  \vspace{-2mm}
  \caption{Comparison of recognition performance with previous methods on STR datasets. MIX represents our generated data and we note its amount in parentheses. The numbers of ``$\Delta$'' in \textcolor{Green}{\textbf{green}} and \textcolor{blue}{\textbf{blue}} denote the improvements over each dataset and average, respectively. }
  \vspace{-1mm}
    \begin{tabular}{cccccccccc}
    \hline
    & \textbf{Methods} & \textbf{Training Dataset} & \textbf{IC13}  & \textbf{SVT}   & \textbf{IIIT}  & \textbf{IC15}  & \textbf{SVTP}  & \textbf{CUTE} & \textbf{Average} \\
    \hline
    & CRNN~\cite{shi2016crnn} & ST    & 86.7  & 80.8  & 78.2  & -     & -     & -   & -  \\
    & TBRA~\cite{Baek2019what} & MJ+ST & 93.6 & 87.5 & 87.9 & 77.6 & 79.2 & 74.0 & 84.6 \\
    & ESIR~\cite{zhan2019esir} & MJ+ST & 91.3 & 90.2 & 93.3 & 76.9 & 79.6 & 83.3 & 87.1 \\ 
    & MORAN~\cite{cluo2019moran} & MJ+ST & 92.4 & 88.3 & 91.2 & 68.8 & 76.1 & 77.4 & 83.3 \\ 
    & ASTER~\cite{shi2018aster} & MJ+ST & 91.8 & 89.5 & 93.4 & 76.1 & 78.5 & 79.5 & 86.4 \\ 
    & SAM~\cite{Liao2021mask} & MJ+ST & 95.3  & 90.6  & 93.9  & 77.3  & 82.2  & 87.8 & 88.3 \\
    & SE-ASTER~\cite{qiao2020seed} & MJ+ST & 92.8  & 89.6  & 93.8  & 80.0    & 81.4  & 83.6 & 88.3 \\
    & TextScanner~\cite{wan2020textscanner} & MJ+ST & 92.9  & 90.1  & 93.9  & 79.4  & 84.3  & 83.3 & 88.5 \\
    & DAN~\cite{DAN_aaai20} & MJ+ST & 93.9  & 89.2  & 94.3  & 74.5  & 80.0    & 84.4 & 87.2 \\
    & RobustScanner~\cite{Yue2020robustscanner} & MJ+ST & 94.8  & 88.1  & 95.3  & 77.1  & 79.5  & 90.3 & 88.4 \\
    & SRN~\cite{Deli2020srn} & MJ+ST & 95.5  & 91.5  & 94.8  & 82.7  & 85.1  & 87.8 & 90.4 \\
    & VisionLAN~\cite{Wang2021visionlan} & MJ+ST &  95.7 &  91.7 &  95.8 &  83.7 & 86.0 & 88.5 & 91.2 \\
    & ABINet~\cite{fang2021abinet} & MJ+ST & 97.4  & 93.5  & 96.2  & 86.0 & 89.3  & 89.2 & 92.6 \\
    \hline
    \multirow{6}[2]{*}{(a)}  & CRNN (Baseline) & ST    & 87.8  & 80.9  & 85.7  & 63.0    & 66.3     & 70.8     & 77.2     \\
     & CRNN (Ours) & ST+MIX (15M) & 89.7  & 81.6  & 87.1  & 65.0  & 67.8     & 75.0     & 78.9     \\
    & $\Delta$ & - & \textcolor{Green}{\textbf{+1.9}} & \textcolor{Green}{\textbf{+0.7}} & \textcolor{Green}{\textbf{+1.4}} & \textcolor{Green}{\textbf{+2.0}} & \textcolor{Green}{\textbf{+1.5}} & \textcolor{Green}{\textbf{+4.2}} & \textcolor{blue}{\textbf{+1.7}}  \\
     & ABINet (Baseline) & MJ+ST    & 97.3  & 93.5  & 96.3  & 86.1    & 89.1     & 89.2     & 92.7     \\
     & ABINet (Ours) & MJ+ST+MIX (30M) & \textbf{98.1}  & \textbf{94.2}  & \textbf{96.6}  & \textbf{86.7}  & \textbf{89.6}     & \textbf{91.7}     & \textbf{93.2}     \\
    & $\Delta$ & - & \textcolor{Green}{\textbf{+0.8}} & \textcolor{Green}{\textbf{+0.7}} & \textcolor{Green}{\textbf{+0.3}} & \textcolor{Green}{\textbf{+0.6}} & \textcolor{Green}{\textbf{+0.5}} & \textcolor{Green}{\textbf{+2.5}} & \textcolor{blue}{\textbf{+0.5}}  \\
    \hline   
    \multirow{6}[2]{*}{(b)} & CRNN (Baseline) & Real-L    & 86.6  & 77.1  & 84.0    & 61.1    & 62.7   & 64.2  & 75.0     \\
     & CRNN (Ours) & Real-L+MIX (2M) & 89.0  & 81.1  & 88.1  & 67.8   & 68.8  & 72.2   & 79.9     \\
    & $\Delta$ & - & \textcolor{Green}{\textbf{+2.4}} & \textcolor{Green}{\textbf{+4.0}} & \textcolor{Green}{\textbf{+4.1}} & \textcolor{Green}{\textbf{+6.7}} & \textcolor{Green}{\textbf{+6.1}} & \textcolor{Green}{\textbf{+8.0}} & \textcolor{blue}{\textbf{+4.9}}  \\
     & ABINet (Baseline) & Real-L    & 95.7  & 93.5  & 96.6    & 86.9    & 86.5   & 94.1  & 92.8     \\
     & ABINet (Ours) & Real-L+MIX (2M) & 97.2  & 96.0  & 97.0  & 89.0   & 90.2  & 95.8   & 94.3     \\
    & $\Delta$ & - & \textcolor{Green}{\textbf{+1.5}} & \textcolor{Green}{\textbf{+2.5}} & \textcolor{Green}{\textbf{+0.4}} & \textcolor{Green}{\textbf{+2.1}} & \textcolor{Green}{\textbf{+3.7}} & \textcolor{Green}{\textbf{+1.7}} & \textcolor{blue}{\textbf{+1.5}}  \\    
    \hline
    \end{tabular}%
  \label{recognition_str}%
  \vspace{-3mm}
\end{table*}%

We compare our method to SOTA methods in Tab.~\ref{recognition_preformance}. 
For a fair comparison, note that methods using additional real data or language models are outside the scope of this study. 
It can be seen that the proposed CTIG-DM outperforms previous methods for data augmentation and data synthesis. 
Moreover, we integrate our method with data augmentation by using the open-source toolkit\footnote{https://github.com/Canjie-Luo/Text-Image-Augmentation} and performing random geometric augmentation~\cite{luo2020learn} on the generated samples. 
Finally, adding data augmentation to our method can further improve the performance, which suggests that the proposed CTIG-DM is complementary to previous works.

\subsubsection{Scene Text Recognition}
Different STR methods may adopt different backbones, data processing, training policies, etc. Therefore, for a fair comparison, we reproduce two representative works CRNN and ABINet to evaluate the effectiveness of our algorithm. 
As shown in Tab.~\ref{recognition_str}~(a), when combined with our generated data, the performance of CRNN is improved by an average of 1.7\% on the six test datasets and that of ABINet is boosted by an average of 0.5\%, which is considerable.

By comparing Tab.~\ref{recognition_preformance} and Tab.~\ref{recognition_str}~(a), we observe that the improvements of CTIG-DM in STR are not as obvious as that of HTR. 
This is probably because most STR methods are trained on synthesized data only, and the training set has about 14M images, which is two orders of magnitude larger. 
Both the diversity and fidelity of the proposed CTIG-DM are partially limited by the synthesized training set. 
However, CTIG-DM can still bring performance improvements to the advanced recognizer without making any changes to the network. 
Further, we evaluate the effectiveness of our method on the real dataset Real-L, whose scale is about 270K. 
As shown in Tab.~\ref{recognition_str}~(b), our generated data brings significant improvements of 4.9\% and 1.5\% to CRNN and ABINet, respectively. 
It proves that the proposed CTIG-DM is able to produce valid image samples that simulate the complexity and diversity of the real world, which can better improve the performance of existing text recognizers.

\begin{table}[htbp]
\setlength\tabcolsep{4pt}
\footnotesize
  \centering
  \caption{Comparison of image generation quality with previous methods on HTR datasets. Lower values are preferable. }
    \begin{tabular}{cccccc}
    \hline
    \multirow{2}[4]{*}{} & \multicolumn{2}{c}{\textbf{RIMES}} &       & \multicolumn{2}{c}{\textbf{IAM}} \\
\cline{2-3}\cline{5-6}          & \textbf{FID}   & \textbf{GS}    &       & \textbf{FID}   & \textbf{GS} \\
    \hline
    Alonso et al.~\cite{alonso2019adver} & 23.94 & $8.58\times10^{-4}$ &       & -     & - \\
    ScrabbleGAN~\cite{fogel2020scrabblegan} & 23.78 & $7.60\times10^{-4}$ &       & 20.72 & $2.56\times10^{-2}$ \\
    HiGAN~\cite{gan2021higan} & -     & -     &       & 17.28 & - \\
    CG-GAN~\cite{kong2022look} & -     & -     &       & 19.03 & - \\
    HWT~\cite{Bhunia2021handt} & -     & -     &       & 19.40  & $1.01\times10^{-2}$ \\
    Davis et al.~\cite{Brian2020text} & 23.72 & $7.19\times10^{-1}$ &       & 20.65 & $4.88\times10^{-2}$ \\
    SLOGAN~\cite{luo2022slogan} & 12.06 & $5.59\times10^{-4}$ &       & -     & - \\
    Ours  & \textbf{9.75}  & $\textbf{6.23}\times\textbf{10}^{-5}$ &       & \textbf{9.34}  & $\textbf{5.91}\times\textbf{10}^{-5}$ \\
    \hline
    \end{tabular}%
  \label{generation_quality}%
  \vspace{-4mm}
\end{table}%

\subsection{Quality of Generated Images}
In this subsection, we compare with previous methods in terms of image generation quality. 
Similar to the settings in~\cite{luo2022slogan}, FID is calculated with 25k real and 25k generated images, while GS is calculated with 5k real and 5k generated images. 
As illustrated in Tab.~\ref{generation_quality}, benefiting from the proposed various condition and powerful generative ability of the diffusion models, our method outperforms the SOTA methods by a notable margin in all metrics on both datasets, which indicates the advancement of CTIG-DM.

\subsection{Generating Out-of-Vocabulary Images}
\label{section_oov}
\begin{figure}[h]
\centering
\includegraphics[width=0.45\textwidth]{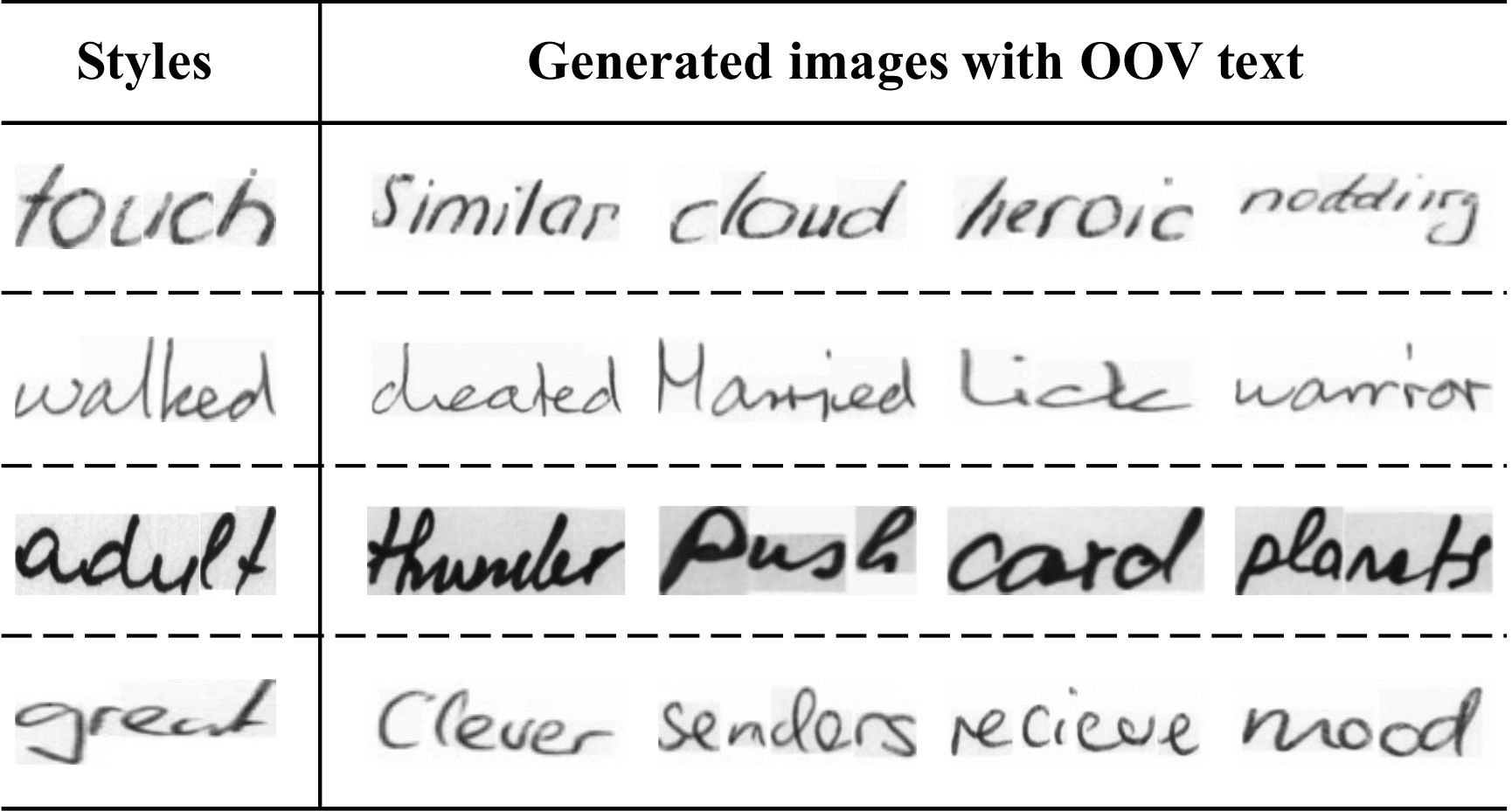}
\vspace{-2mm}
\caption{Visualization of images generated by CTIG-DM according to the specified styles and OOV text. Note that none of the words on the right appear in the training set. } \label{figure_oov}
\vspace{-3mm}
\end{figure}

As described in Sec.~\ref{generation_process}, the proposed synthesis mode only depends on the text condition. 
Therefore, by changing the input text, we can generate images containing OOV text. 
In this subsection, we explore the OOV image generation quality of the proposed CTIG-DM. 
Since previous works~\cite{Kang2020ganwriting,luo2022slogan,kong2022look,Bhunia2021handt} generate OOV images with the specified styles, for a fair comparison, we use text condition and style condition as the input conditions.
As presented in Tab.~\ref{oov_experiment}, compared with previous methods, the proposed CTIG-DM significantly boosts the OOV image generation quality. 
Fig.~\ref{figure_oov} shows the visualization of images generated by CTIG-DM, which suggests the controllability of our method over the styles and text of the generated images.

\begin{table}[htbp]
  \centering
  \vspace{-2mm}
  \caption{Comparison of OOV image generation quality with previous methods on the IAM dataset. }
  \vspace{-2mm}
    \begin{tabular}{cc}
    \hline
    \textbf{Method} & \textbf{FID} \\
    \hline
    GANwriting~\cite{Kang2020ganwriting} & 125.87 \\
    HWT~\cite{Bhunia2021handt}  & 109.45 \\
    SLOGAN~\cite{luo2022slogan} & 97.81 \\
    CG-GAN~\cite{kong2022look} & 104.81 \\
    Ours  & \textbf{25.52} \\
    \hline
    \end{tabular}%
  \label{oov_experiment}%
  \vspace{-3mm}
\end{table}%

\subsection{Domain Adaptation}

\begin{table}[htbp]
\setlength\tabcolsep{4pt}
  \centering
  \vspace{-3mm}
  \caption{Comparison of domain adaptation capacity with previous methods on the CVL dataset. SYN represents our generated data using the lexicon of the CVL training set. }
  \vspace{-2mm}
    \begin{tabular}{cccc}
    \hline
    \textbf{Method} & \textbf{Training Data} & \textbf{WER}   & \textbf{CER} \\
    \hline
    Baseline & IAM   & 42.64 & 18.49 \\
    ScrabbleGAN~\cite{fogel2020scrabblegan} & IAM+GAN (100K) & 35.98 & 17.27 \\
    SLOGAN~\cite{luo2022slogan} & IAM+GAN (100K) & 34.98 & 14.10 \\
    Ours  & IAM+SYN (100K) & \textbf{26.24} & \textbf{10.89} \\
    \hline
    \end{tabular}%
  \label{domain_adaption}%
  \vspace{-2mm}
\end{table}%

In this subsection, we investigate the domain adaptation capacity of our method. 
Following the settings of the previous works~\cite{fogel2020scrabblegan,luo2022slogan}, we first train CTIG-DM with the IAM training set and then generate 100K samples using the lexicon of the CVL training set. 
Finally, combining the IAM training set and generated samples to train the recognizer, we evaluate the performance on the test set of CVL. 
We repeat the training five times and report the averages. 
As illustrated in Tab.~\ref{domain_adaption}, the performance improvements brought by our generated data far exceed that of the previous methods. 
Concretely, compared with the baseline, WER and CER are decreased by 16.40\% and 7.60\%, respectively, indicating the superior validity and diversity of our generated data. 

\begin{figure}[h]
\centering
\vspace{-1mm}
\includegraphics[width=0.48\textwidth]{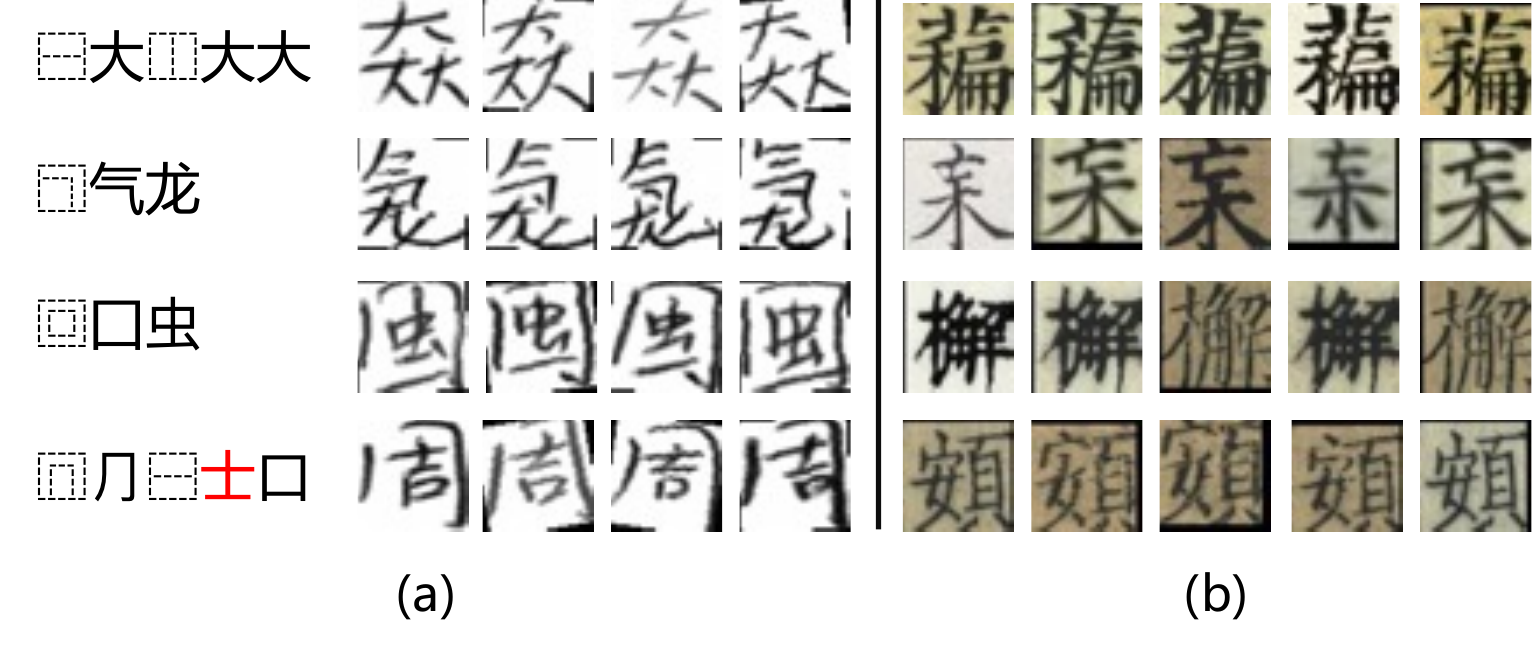}
\vspace{-6mm}
\caption{Visualization of (a) non-existing handwritten Chinese characters (radical sequences are presented in the first column) and (b) ancient rare characters.}
\label{figure_application}
\vspace{-5mm}
\end{figure}

\subsection{Applications} \label{applications}

The above experiments already show that CTIG-DM can advance Latin text image generation. In this subsection, we explore the applications of CTIG-DM in other scenarios, e.g., the generation of OOV handwritten Chinese and ancient characters, to illustrate the potential of our method.

\textbf{OOV handwritten Chinese characters generation.}
Compared with Latin characters such as English and French, Chinese characters are more complex in structure and contain more information. 
Therefore, in order to make full use of the Chinese character prior, we decompose the Chinese characters into radical sequences as text conditions. 
Similar to generating OOV words in English, we can construct radical sequences that are not in the training set to generate OOV Chinese characters or create non-existing handwritten Chinese characters. 
Fig.~\ref{figure_application} (a) shows some visualizations where the characters are not in the training set or are even nonexistent. 
Specifically, the characters shown in the last line are typos, which may come into play in scenarios where typo recognition is required. 

\textbf{Ancient characters generation.} 
Ancient text recognition has great significance in the inheritance of historical knowledge, literature, and art. 
One of its main difficulties is the long-tailed distribution, i.e., it lacks training samples for rare characters. 
Therefore, we can use CTIG-DM to generate rare characters (shown in Fig.~\ref{figure_application} (b)). 
In addition, since rare characters generally have an extremely small number of samples, the method of combining CTIG-DM and radicals for characters generation is also promising. 

\section{Conclusion and Future Work}
We have presented a text image generation method, called CTIG-DM, based on diffusion models. To adapt to the nature of text images, we devise three conditions and four generation modes. Extensive experiments have demonstrated the effectiveness and advantage of CTIG-DM. Specifically, it can be used to boost the performance of existing text recognizers. 
In the future, we will explore more generation modes on more types of text images, and generate text images with the styles of unseen writers.

{\small
\bibliographystyle{ieee_fullname}
\bibliography{egbib}
}

\end{document}